\pdfoutput=1
\documentclass[10pt,twocolumn,letterpaper]{article}

\usepackage{cvpr}
\usepackage{times}
\usepackage{epsfig}
\usepackage{graphicx}
\usepackage{amsmath}
\usepackage{amssymb}
\usepackage{subcaption}
\usepackage{caption}
\usepackage{xcolor}
\usepackage[font=small,labelfont=bf,tableposition=top]{caption}
\usepackage{booktabs}
\usepackage{multirow}
\usepackage{algorithm,algorithmic}
\usepackage[numbers,sort]{natbib} 
\usepackage{authblk}
\usepackage{xspace}
\usepackage{enumitem}
\usepackage{pifont}
\usepackage{dsfont}

\usepackage{hyperref}
\hypersetup{pagebackref=true,breaklinks=true,colorlinks,bookmarks=false}

\cvprfinalcopy

\ifcvprfinal\fi
\begin{document}
\newcommand{\repnet}{RepNet\xspace}
\newcommand{\countix}{Countix\xspace}
\newcommand{\tr}{^{^\intercal}} 

\makeatletter
\renewcommand\AB@affilsepx{  \protect\Affilfont}
\renewcommand\thesubfigure{\thefigure\alph{subfigure}} 
\renewcommand\p@subfigure{}
\makeatother

\title{
A Short Note on Evaluating RepNet for Temporal Repetition Counting in Videos
}

\author[ 1]{Debidatta Dwibedi}
\author[ 1]{Yusuf Aytar}
\author[ 1]{Jonathan Tompson}
\author[ 1]{Pierre Sermanet}
\author[ 1]{Andrew Zisserman}
\affil[ 1 ]{Google DeepMind 

{\small \tt \{debidatta, yusufaytar, tompson, sermanet, zisserman\}@google.com}}

\maketitle

\begin{abstract}
We discuss some consistent issues on how RepNet has been evaluated in various papers. As a way to mitigate these issues, we report RepNet performance results on different datasets, and release evaluation  code and the RepNet checkpoint to obtain these results. Code URL: \url{https://github.com/google-research/google-research/blob/master/repnet/}
\end{abstract}

\section{Introduction}

This note is related to evaluating the class-agnostic repetition counting model RepNet~\cite{dwibedi2020counting} on various video repetition counting datasets.

In many papers~\cite{hu2022transrac, sinha2024every,xinjieracmfl24,luo2024rethinking} it has been reported that performance of RepNet on some repetition counting datasets is deficient. The first time this was reported was in the TransRAC~\cite{hu2022transrac} paper. However, the model referred to as `RepNet' in that paper is a modified version of RepNet. In Section 5.3 of~\cite{hu2022transrac}, it is mentioned that \textit{"... for a fair comparison, we modify the last fully connection layer of RepNet~\cite{dwibedi2020counting} to make it capable of handling those videos containing more than 32 action periods"}. It is unclear what this modification is exactly but it leads to the modified RepNet's performance being close to 0 in the Off-by-One Accuracy (OBOA) metric on the  UCFRep and RepCount datasets. These results imply that the modified model is not able to count repetitions in the videos of these datasets. The papers that follow TransRAC have reused the numbers reported in TransRAC but refer to this modified model as RepNet. 

We would like to highlight that the original RepNet model {\em is} capable of making predictions of higher than 32 period length. This is achieved by playing the video at different speeds, rather than modifying the model. This technique is described in the original RepNet paper~\cite{dwibedi2020counting} as \textit{Multi-speed Evaluation} (Section 3.5). This was used for evaluating on the Countix dataset as well and has been proposed before in~\cite{levy2015live}. When we evaluate the RepNet model using the multi-speed technique on the UCFRep and RepCount datasets, we find that it results in strong performance. 

\section{Evaluation Results}

We first define the evaluation metrics, and then report the performance
of the original RepNet model on the following datasets: Countix, UCFRep and RepCount-A. 

\subsection{Evaluation metrics}
Consider an evaluation dataset consisting of $N$ videos where 
$c_i$ is the ground truth count of the $i^{th}$ video and $\tilde{c}_i$ is the predicted count of the $i^{th}$ video. Using the above, we define the commonly used metrics as follows: 

\noindent \textbf{Off-by-one Accuracy (OBOA):} 

\begin{minipage}{\linewidth}
\begin{equation}
  OBOA = \frac{1}{N}\sum_{i \in N} \mathds{1}(| c_i - \tilde{c}_i | \leq 1 ) ,
  \label{eqn:OBO}
\end{equation}
\end{minipage}

Note that the original RepNet paper~\cite{dwibedi2020counting} reports OBOE (off-by-one error) which is mathematically equal to $1.0\,-\,$OBOA.

\noindent \textbf{Mean Absolute Error (MAE):} 

\begin{minipage}{\linewidth}
\begin{align}
  MAE = \frac{1}{N}\sum_{i \in N} \frac{| c_i - \tilde{c}_i|}{(\alpha + c_i)} ,
  \label{eqn:MAE}
\end{align}
\end{minipage}

In RepNet~\cite{dwibedi2020counting} $\alpha=0$ is used. However, the  TransRAC codebase~\cite{hu2022transrac} uses $\alpha=0.1$ ~\cite{transraccodebase}. This usage of $\alpha=0.1$ has been followed by ~\cite{sinha2024every} in their implementation as well~\cite{esccodebase}. 
This is to ensure that a divide by 0 problem does not happen for videos without any repetition. However, this is different from the metric used in earlier papers~\cite{levy2015live,runia2018real,dwibedi2020counting}. We advise caution to others using this metric to keep the extra 0.1 in mind while reporting their performance to have a fair comparison with other methods. We explicitly mention the value of $\alpha$ when reporting the metric in this paper.

\subsection{RepNet counting performance}

\begin{table*}[!t]
\centering
\normalsize{
    \begin{tabular}{lc|c|lll}
       \textbf{\hspace{1.5cm} Model} & \textbf{Year} & \textbf{MAE $\alpha$} & \textbf{MAE $\downarrow$}& \textbf{OBOA $\uparrow$} & \textbf{OBOE $\downarrow$}\\
        \midrule
        
        \repnet (from paper) & 2020 & 0.0 & 0.364 & 0.697 & 0.303\\
        \midrule
        RACNet \cite{luo2024rethinking} & 2024 & 0.0 & 0.5278 & 0.3924 & 0.6076\\
         \repnet  & 2020 & 0.0 & \textbf{0.3083} & \textbf{0.7047} & \textbf{0.2953}\\
         \midrule
        ESCounts \cite{sinha2024every} & 2024 & 0.1 & \textbf{0.276} & 0.673 & 0.327\\
         \repnet  & 2020 & 0.0 & 0.3002 & \textbf{0.7047} & \textbf{0.2953}\\
    \bottomrule
    \end{tabular}
\caption{Counting Results on the {\bf Countix dataset} (test split). Note, 
OBOE (off-by-one error) is equal to $1 - $OBOA (off-by-one accuracy).
}
\label{tab:countix_results}
\vspace{-2.5em}
}
\end{table*}

\noindent\textbf{Countix.}  For the repetition counting task in the Countix dataset, the videos are segmented out into the start and end of the entire repeating segment. The task of the model is to count the repetitions in this segment. As described in~\cite{dwibedi2020counting}, we evaluate videos using RepNet at different frame rates. We
play the video at 1$\times$, 2$\times$, 3$\times$,  4$\times$, 5$\times$ speeds. This multi-speed evaluation technique allows us to use the same model to handle periods of lengths longer than 32 frames.
By sampling at a maximum speed of 5$\times$ we are able to increase the maximum period length that can be predicted by RepNet to 160 frames. We report the results on the Countix dataset~\cite{dwibedi2020counting} in Table~\ref{tab:countix_results}. We find that RepNet outperforms more modern methods such as~\cite{sinha2024every}. Please note that we are reporting Off-by-One Accuracy here as opposed to Off-by-One Error in the original RepNet paper as most papers report accuracy on these benchmarks now. 

\noindent\textbf{UCFRep.} UCFRep was introduced in~\cite{Zhang_2020_CVPR}. This dataset is similar to Countix. Hence, we use the same evaluation settings as Countix.
We present the results in Table~\ref{tab:ucfrep_results}. We not only find that RepNet outperforms TransRAC significantly but also it is closer in performance to more modern methods that use much larger backbone networks at higher resolution for testing. 

\begin{table}[!h]
\vspace{-.5em}
\centering
\normalsize{
    \begin{tabular}{l|c|l|l}
       \textbf{\hspace{1.5cm} Model} & \textbf{Year} & \textbf{MAE $\downarrow$ }& \textbf{OBOA $\uparrow$}\\
        \midrule
        Mod.RepNet in TransRAC \cite{hu2022transrac} & 2022 & 0.9985 & 0.009\\
        \midrule
        TransRAC \cite{hu2022transrac} & 2022 & 0.6401 & 0.324\\
        RACNet $^*$ \cite{luo2024rethinking}   & 2024 & 0.5260 & 0.3714\\
        MFL $^*$ \cite{xinjieracmfl24}  & 2024 &  0.388 & 0.510 \\
        ESCounts  \cite{sinha2024every} & 2024 & 0.216 & 0.704\\
        \repnet & 2020 & \textbf{0.2088} & \textbf{0.7333}\\
    \bottomrule
    \end{tabular}
\caption{Counting Results on the {\bf UCFRep dataset} (val split). MAE is reported with $\alpha=0.1$. $^*$ denotes that we were not able to verify the value of $\alpha$. }
\label{tab:ucfrep_results}
\vspace{-1.5em}
}
\end{table}

\noindent\textbf{RepCount-A.} This dataset was introduced in~\cite{hu2022transrac}. The key factor that differentiates RepCount-A from Countix and UCFRep is that RepCount-A videos can contain gaps in repetitions. That is, the model has to predict both the repeating segment as well as the count within the segment. RepNet can handle such videos because it is trained to predict both per-frame periodicities and per-frame period lengths. To evaluate in this setting, we use both the per-frame periodicity score and confidence of predicted period to determine whether a particular frame is to be counted as repeating or not. Concretely, we sample consecutive non-overlapping windows of $N$ frames and provide it as input to \repnet which outputs per-frame periodicity $p_i$ and period lengths $l_i$. We found using both periodicity predictions $p_i$ and period length prediction scores $l_i$ to give a more accurate measure of whether a frame should be considered repeating or not.  We define \textit{per-frame predicted count} $\tilde{c}_i$ as follows:

\begin{minipage}{\linewidth}
\begin{equation}
  \tilde{c}_i =\frac{\mathds{1}({\sqrt{(p_i \times l^s_i)}}> \tau)}{l_i}
  \label{eqn:count_formula}
\end{equation}
\end{minipage}

where $\tau$ is a binary threshold on the periodicity score. We use $\tau=0.5$ in our experiments. The overall repetition count is computed as the sum of all per-frame counts: $\tilde{c} = \sum_{i=1}^{N} \tilde{c}_i $. As done in previous sections, we use multi-speed evaluation for RepCount-A dataset as well. We play the video at 1$\times$, 2$\times$, 3$\times$,  4$\times$ and 5$\times$ speeds and choose the speed with the highest period length score.

We present the results in Table~\ref{tab:repcount_results}. First, we find that RepNet still performs well for videos with gaps in repetitions. This is contrary to what has been reported in the TransRAC paper (Row 1) and follow-up papers that use the ``Modified RepNet" baseline reported in the TransRAC paper. The only model outperforming RepNet is ESCounts which has been trained on RepCount-A and tests at a resolution of $224\times224$. In comparison, RepNet has not been trained on RepCount-A and processes videos at a resolution of $112\times112$.

\begin{table}[!h]
\vspace{-.5em}
\centering
\normalsize{
    \begin{tabular}{l|c|l|l}
       \textbf{\hspace{1.5cm} Model} & \textbf{Year} & \textbf{MAE $\downarrow$ } & \textbf{OBOA $\uparrow$ }\\
        \midrule
        Mod. RepNet in TransRAC \cite{hu2022transrac} & 2022 & 0.9950 & 0.0134\\
        \midrule
        TransRAC \cite{hu2022transrac} & 2022 & 0.4431 & 0.2913\\
        MFL$^*$ \cite{xinjieracmfl24} &2024 &  0.384 & 0.386 \\
        RACNet$^*$ \cite{luo2024rethinking} & 2024 &0.4441 & 0.3933 \\
        ESCounts \cite{sinha2024every} &  2024 & \textbf{0.245} & \textbf{0.563}  \\
        \repnet & 2020 & 0.3308 & 0.5329 \\
    \end{tabular}
\caption{Counting Results on the {\bf RepCount-A dataset} (test split). MAE is reported with $\alpha=0.1$. $^*$ denotes that we were not able to verify the value of $\alpha$.}
\label{tab:repcount_results}
\vspace{-3.em}
}
\end{table}

\section{Conclusion}
We have released the code and checkpoints for RepNet. We hope this will clear up any confusion about RepNet evaluation. However, this re-evaluation raises the question of what modern methods and larger backbones have brought to video repetition counting methods. In particular, a 2020 model trained with ResNet-50 backbone evaluated at $112\times112$ resolution still has strong performance. We hope that the community can use the released models to further improve repetition counting methods.

{\small
\bibliographystyle{ieee_fullname}
\bibliography{egbib}
}

\end{document}